\documentclass[twocolumn]{article}
\usepackage{cvpr}              

\usepackage{graphicx}
\usepackage{amsmath}
\usepackage{amssymb}
\usepackage{booktabs}
\usepackage{float}
\usepackage{multirow}
\usepackage{booktabs}
\usepackage{setspace}
\usepackage{color}
\usepackage{algorithm}
\usepackage{algpseudocode}
\usepackage{algorithmicx}
\usepackage{threeparttable}

\usepackage{xurl}
\usepackage{bbding}
\usepackage[pagebackref,breaklinks,colorlinks]{hyperref}

\usepackage[capitalize]{cleveref}
\crefname{section}{Sec.}{Secs.}
\Crefname{section}{Section}{Sections}
\Crefname{table}{Table}{Tables}
\crefname{table}{Tab.}{Tabs.}

\def\cvprPaperID{4525} 
\def\confName{CVPR}
\def\confYear{2022}

\begin{document}

\title{Scene Consistency Representation Learning for Video Scene Segmentation}

\author{Haoqian Wu$^{1,2,3,4*}$, Keyu Chen$^{2*}$, Yanan Luo$^{2}$, Ruizhi Qiao$^{2}$, Bo Ren$^{2}$, \\ 
Haozhe Liu$^{1,3,4,5}$, Weicheng Xie$^{1,3,4}{}^\dagger$, Linlin Shen$^{1,3,4}$ \\
$^1$ Computer Vision Institute, Shenzhen University $^2$ Tencent YouTu Lab \\
$^3$ Shenzhen Institute of Artificial Intelligence and Robotics for Society \\
$^4$ Guangdong Key Laboratory of Intelligent Information Processing $^5$ KAUST \\
{\tt\small wuhaoqian2019@email.szu.edu.cn}
{\tt\small \{yolochen, ruizhiqiao, timren\}@tencent.com} \\ 
{\tt\small luoyanan93@gmail.com} 
{\tt\small haozhe.liu@kaust.edu.sa}
{\tt\small \{wcxie, llshen\}@szu.edu.cn}
}

\twocolumn[{%
\maketitle
\begin{figure}[H]
\vspace{-3.0em}
\hsize=\textwidth 
\centering
\includegraphics[width=.90\textwidth]{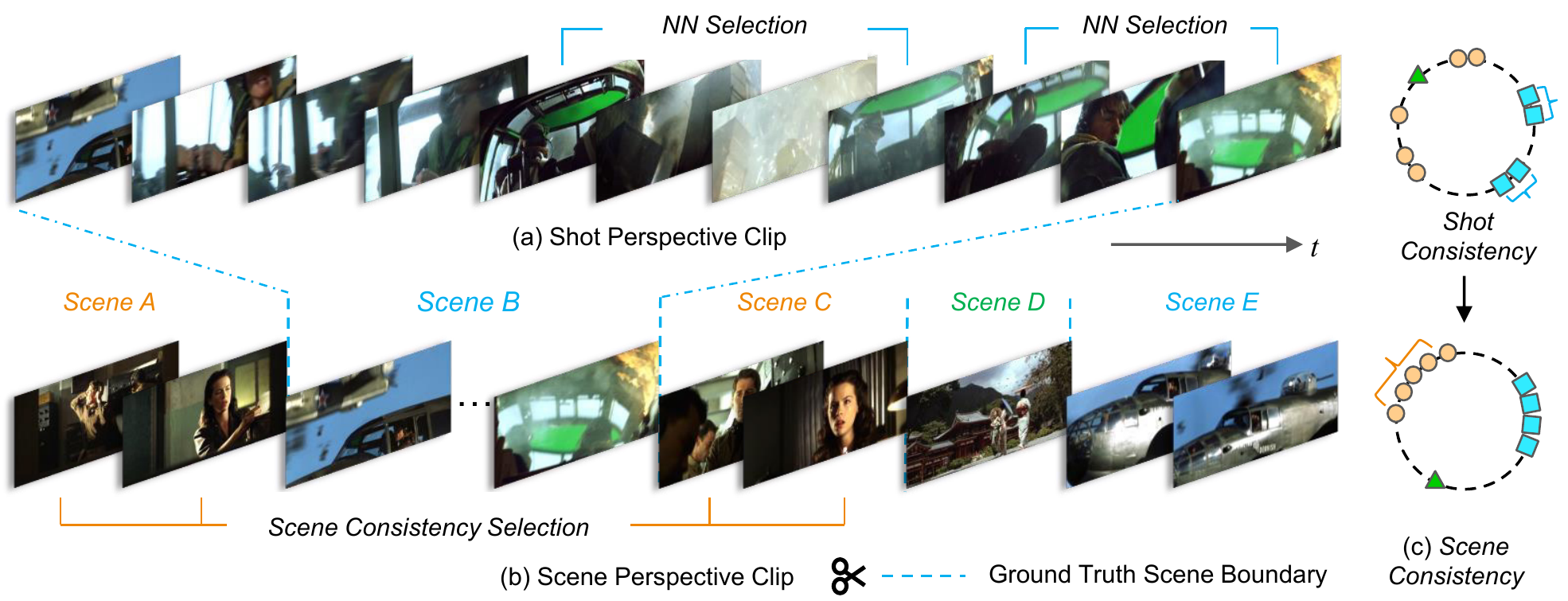}
\caption{\textbf{An illustration of representation learning methods from the shot-to-scene perspective}. Several continuous shots are shown in Fig. (a), where existing SSL approaches obtain positive pairs from the adjacent shots (\textit{e.g}., by performing \textit{Nearest Neighbor (NN) Selection} \cite{chen2021shot}). While we propose to look further for scenes that are often crossed over, as \textcolor[RGB]{239,134,0}{\textit{Scene A/C}} and \textcolor[RGB]{0,176,240}{\textit{Scene B/E}} shown in Fig. (b), where positive samples are explored in a broader region and the shots are clustered to the same scene in the feature representation space, \textit{i.e.}, Fig. (c). Best viewed in color.} 
\label{fig:problem}
\end{figure}
}]

\footnotetext{\noindent$^*$Equal Contribution \quad $^\dagger$Corresponding Author}

\begin{abstract}
    A long-term video, such as a movie or TV show, is composed of various scenes, each of which represents a series of shots sharing the same semantic story. Spotting the correct scene boundary from the long-term video is a challenging task, since a model must understand the storyline of the video to figure out where a scene starts and ends. 
    To this end, we propose an effective Self-Supervised Learning (SSL) framework to learn better shot representations from unlabeled long-term videos. More specifically, we present an SSL scheme to achieve scene consistency, while exploring considerable data augmentation and shuffling methods to boost the model generalizability. 
    Instead of explicitly learning the scene boundary features as in the previous methods, we introduce a vanilla temporal model with less inductive bias to verify the quality of the shot features. Our method achieves the state-of-the-art performance on the task of Video Scene Segmentation. Additionally, we suggest a more fair and reasonable benchmark to evaluate the performance of \textit{Video Scene Segmentation} methods. The code is made available.\textcolor{red}{$^1$} 
\end{abstract}

\footnotetext{$^1$\url{https://github.com/TencentYoutuResearch/SceneSegmentation-SCRL}}

\section{Introduction}
\label{sec:intro}

In the process of video creation, to make the story more compelling, the editor will use various editing techniques, such as montage, one shot to the end, \textit{etc.} Quickly switching between stories and scenes makes the movie plot tighter, \textit{e.g.} inserting outdoor battle scenes into indoor dialogue scenes, as shown in Fig. \ref{fig:problem} (b), making the scene transition more intriguing and unpredictable, thus the task of \textit{Video Scene Segmentation} turns out to be rather challenging.
Hence, it is essential to understand the high-level semantic information of each scene in the long-term video.


There has been extensive studies dealing with video understanding tasks on datasets where the individual video clip is typically short, while requiring a lot of labor to segment uncurated videos into short videos by category. Although some studies focus on splitting the long video into smaller segments, \textit{e.g.}, the methods of \textit{Action Spotting} \cite{lin2018bsn,tan2021relaxed,cioppa2020context,giancola2018soccernet} aim to locate the positions of the beginning and ending of the action, however, they are the category-aware approaches. By contrast, \textit{Video Scene Segmentation} is a category-agnostic task that only the scene boundary label is available, and it’s very confusing to classify a scene fragment taxonomically.

Since a long-term video is inherently structured in a specific way, a sequence of frames can be divided into \textbf{shots} or \textbf{scenes} in terms of the granularity of semantics \cite{rao2020local} \cite{sidiropoulos2011temporal}. More specifically, a \textbf{shot} contains only continuous frames taken by the camera without interruption, and a \textbf{scene} is composed of successive shots and describes the same short story. For detecting shot  boundaries, \cite{lokovc2019framework} \cite{sidiropoulos2011temporal} split a video into many separate shots using lower-level visual context. Based on this, many mainstream approaches of \textit{Video Scene Segmentation} \cite{chasanis2008scene} \cite{ baraldi2015deep} \cite{ rao2020local} \cite{chen2021shot} determine scene boundaries by exploring semantic correlations among the adjacent shots.

While computer vision tasks suffer from the high cost of manual annotation, Self-Supervised Learning (SSL) based methods \cite{he2020momentum,chen2020simple,chen2020big,chen2020improved,chen2021empirical,chen2021exploring,grill2020bootstrap,caron2020unsupervised} are proposed to train a general feature extractor using unlabeled data. 
By leveraging a small amount of annotated data for training, these SSL methods can achieve appealing feature representation to even rival some supervised learning methods.
For \textit{Video Scene Segmentation}, \cite{chen2021shot} proposes to narrow the feature representation distance of the most similar shot pair in a local region, it significantly surpasses the supervised learning method \cite{rao2020local} by employing a mere MLP classifier. 
However, in current SSL methods on the task of \textit{Video Scene Segmentation}, the strategy of positive sample selection, pretraining protocol, evaluation metric and downstream model are not well discussed or addressed.

To achieve this goal, we propose a self-supervised learning 
scheme to learn better representations, as well as the evaluation metric for the task of \textit{Video Scene Segmentation}. The contributions of this paper are summarized as follows:
\begin{itemize}
  \item A representation learning scheme based on Scene Consistency is proposed to obtain better shot representations on the unlabeled long-term video.
  \item A simple yet effective temporal model with less inductive bias is proposed to assess the quality of the shot representation for the downstream \textit{Video Scene Segmentation} task.
  \item A benchmark that is more fair and reasonable is introduced for both pretraining and evaluation. 
  More importantly, the proposed method outperforms the state-of-the-art methods under all the protocols, and can significantly improve the performance of existing supervised methods without bells and whistles.
\end{itemize}


\begin{figure*}[htbp]
  \centering
\includegraphics[width=.92\textwidth]{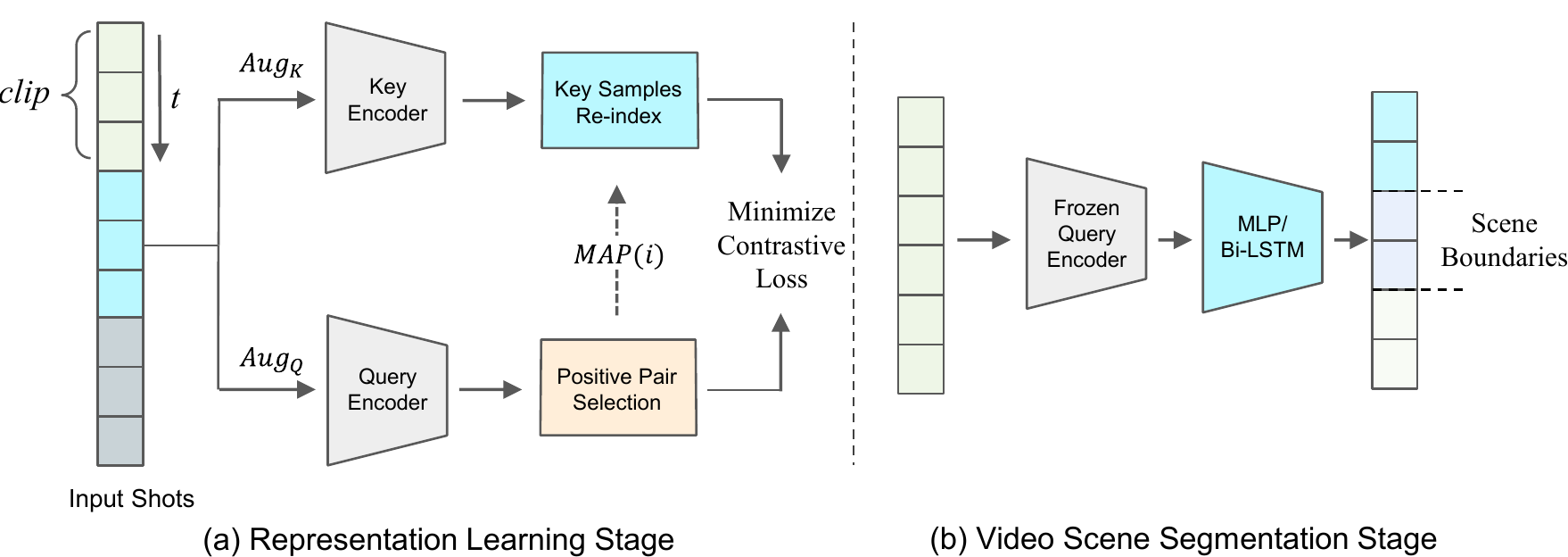}
  \caption{\textbf{The pipeline of the proposed method}. (a) Unsupervised Representation Learning Stage for learning shot representations, where $Map(i)$ is the mapping function for selecting positive samples. (b) Supervised \textit{Video Scene Segmentation} Stage, where the quality of the shot representations is evaluated under the protocols of the non-temporal (MLP) and temporal (Bi-LSTM \cite{huang2015bidirectional}) models.}
  \label{fig:ppl}
  \vspace{-0.8em}
\end{figure*}

\section{Related Work}

\textbf{Self-Supervised Learning in Images and Videos.}
To address the problems of the insufficient and expensive manual annotation, many approaches explore the inherent knowledge in unlabeled data by designing a lot of pretext tasks, including predicting the transformations of images, \textit{e.g.}, image rotation \cite{gidaris2018unsupervised}, inpainting \cite{pathak2016context}, colorizing \cite{zhang2016colorful}, jigsaw \cite{noroozi2016unsupervised}, etc. 
In short, these Self-Supervised Learning (SSL) methods use the information explored from the data themselves for the supervision.
Recently, \cite{he2020momentum,chen2020simple,chen2020big,chen2020improved,chen2021empirical,chen2021exploring,grill2020bootstrap,caron2020unsupervised} introduce the contrastive  similarity metrics to learn invariant feature representation of various views augmented from the original image,
where strong data augmentations \cite{chen2020big} are frequently used in image-level SSL methods to improve the robustness of the learned representations.
From another aspect, by finetuning the model with a small amount of labeled data, SSL methods can achieve competitive performance compared with supervised learning methods, furthermore, the pretrained model can be used in specific downstream tasks.
For video-oriented SSL methods, \cite{qian2021spatiotemporal,diba2021vi2clr,kuang2021video,feichtenhofer2021large,xie2021unsupervised,miech2020end,recasens2021broaden} show the appealing performance and potential on the task of video classification, while their positive pairs are selected from the adjacent clips within a same video. Meanwhile, most of studies are based on short videos and the quality of learned features is assessed  based on video classification. 
Hence, it is meaningful to explore a suitable SSL scheme for tasks with long-term videos.

\textbf{Video Shot Boundary Detection and Scene Segmentation.}
For \textit{Video Scene Segmentation}, shot boundary detection is often conducted in advance, which is specified as a task of locating the transition positions in videos based on the similarity of the frames.
3D convolutional networks and color histogram differencing \cite{lokovc2019framework} are used to identify the transition boundaries.
Based on the shot boundaries, \cite{rao2020local} learns the local and global shot representations and utilizes them to split the continuous shots into scenes according to the transition of the story. 
More specifically, identification of each shot’s segmentation point is treated as a binary classification, which is free to the location of the shot. \cite{chen2021shot} leverages unlabeled video data to obtain shot representations, which outperforms many supervised learning methods on the downstream task of \textit{Video Scene Segmentation}. However, this method is pretrained on the entire video data of MovieNet \cite{huang2020movienet} that include the testing videos, \textit{i.e.} the training protocol is inconsistent with that of conventional Self-Supervised Learning methods \cite{qian2021spatiotemporal} \cite{diba2021vi2clr}.
For evaluating \textit{Video Scene Segmentation} approaches, the datasets of OVSD \cite{rotman2017optimal}, BBC planet earth \cite{baraldi2015deep}, MovieNet \cite{huang2020movienet} and AdCuepoints \cite{chen2021shot} are frequently employed. 

In this work, we propose an unsupervised representation learning method based on scene consistency and a reasonable evaluation scheme for \textit{Video Scene Segmentation} task. 


\section{Methodology}
As shown in Fig. \ref{fig:ppl}, we aim to obtain scene consistency representations on unlabeled long-term videos and design a more reasonable benchmark to verify the quality of the extracted features on the task of \textit{Video Scene Segmentation}. To this end, we
(i) propose a Self-Supervised Learning scheme based on a novel non-temporal selection strategy to achieve scene consistency from various shots, 
and (ii) introduce a vanilla temporal model with less inductive bias as well as the corresponding benchmark for this segmentation task.

\subsection{Consistency based Representation Learning}

Approaches of Self-Supervised Learning (SSL) aim to model representation consistency to enhance network robustness against various variations, \textit{e.g.} spatial or temporal transformations.
In this work, we use an SSL framework of Siamese network to achieve the representation consistency.

More precisely, for a given query shot, the objective is to
(i) maximize the similarity between the representations of query shot and positive samples, \textit{i.e.}, key shots;
(ii) minimize the similarity of the negative sample pairs if they exist.
As shown in Fig. \ref{fig:ppl} (a), the input samples $X$ are first augmented, \textit{i.e.}, $Q = Aug_Q(X), K = Aug_{K}(X) $, and the $i$-th positive pair $\{q, k^{+}\}$ is formulated as follows: 
\begin{equation}
\begin{aligned}
  \{q, k^{+}\} = \{f\left(Q[i] \mid \theta_{Q}\right),\ f\left(K[MAP(i)] \mid \theta_{K}\right)^{+}\} 
\end{aligned}
  \label{eq:aug}
\end{equation}
where $[\cdot ]$ stands for the indexing operation, $f\left(\cdot \mid \theta_{Q}\right)$ and $f\left(\cdot \mid \theta_{K}\right)$ are the encoders with parameters $\theta_{Q}$ and $\theta_{K}$, respectively, 
$MAP(i)$ is the mapping function for selecting positive samples.

\begin{figure}[H]
  \centering
  \includegraphics[width=.42\textwidth]{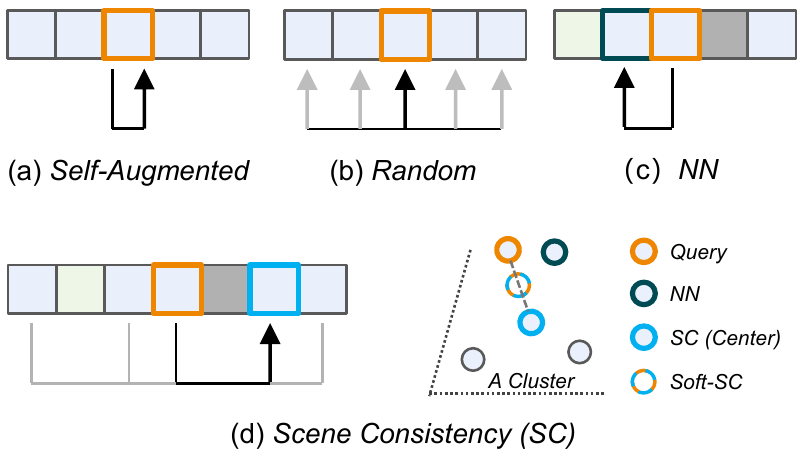}
  \caption{The illustration of four different selection strategies for positive pairs. Best viewed in color.}
  \label{fig:positive}
\end{figure}

For the selection of positive samples in SSL methods based on video data,  three selection strategies are frequently employed, 
\textit{i.e.}, \textit{Self-Augmented} \cite{chen2020improved}, \textit{Random} \cite{feichtenhofer2021large} and \textit{Nearest Neighbor (NN)} \cite{chen2021shot} selections.
For clarity, the three conventional selection strategies for positive samples are represented in Fig. \ref{fig:positive} (a)-(c).

\subsubsection{Conventional Positive Sample Selections}
\label{sec:pos}

\textbf{Self-Augmented Selection.} 
As image-level SSL approaches, the augmented view of one shot is frequently used as its positive sample, as shown in Fig. \ref{fig:positive} (a), the mapping function, \textit{i.e.} the identity mapping, is employed as follows:
\begin{equation}
\begin{aligned}
MAP_{SA}(i) = i
\end{aligned}
\label{eq:id}
\end{equation}

\textbf{Random Selection.} As some SSL methods \cite{feichtenhofer2021large} \cite{qian2021spatiotemporal} for video classification, we select two adjacent shots of the same video as the positive pair, as shown in Fig. \ref{fig:positive} (b), and the mapping function can be formulated as follows:
\begin{equation}
\begin{aligned}
MAP_{RS}(i) = max(i + j,0) 
\end{aligned}
\label{eq:random}
\end{equation}
where $\ j \in \{-n,-n+1,\ ...\ ,n-1,n\}$ and $n$ denotes the size of the search region around the  $i$-th shot.

\textbf{Nearest Neighborhood (\textit{\textbf{NN}}) Selection.} As shown in Fig. \ref{fig:positive} (c), \cite{chen2021shot} proposed to select the positive shot with the closest representation distance to the query shot within a fixed range, and the mapping function is as follows:
\begin{small}
\begin{equation}
\begin{aligned}
  \label{eq:nn}
  MAP_{NN}(i) = \arg\max _{j \in I_{M}}f\left(Q[i] \mid \theta_{Q}\right) \cdot f\left(Q[max(j,0)] \mid \theta_{Q}\right) 
\end{aligned}
\end{equation}
\end{small}
where $I_{M} = \{i-m, \ ...\ , \ i-1,\ i+1, \ ... \ , i+m \} $, $ I_{M}$ stands for the indices of candidate samples for \textit{NN} selection, $m$ is the \textit{search region size} of a given shot, and $2m+1$ is the length of the sliding window.

\subsubsection{Scene Consistency Selection}

\label{sec:SC}
In this work, we propose the \textit{Scene Consistency} Selection, while exploring considerable data augmentation and shuffling methods for the task of \textit{Video Scene Segmentation}.

\textbf{Positive Sample Selection with Scene Consistency.}
As shown in Fig. \ref{fig:problem}, for the video with non-linear narrative, previous selection methods may not work in the case that the most matching shots are far away. Therefore, 
we propose to select positive shot pair based on scene consistency, while the main advantage over \textit{Random/NN} Selection is that our method is non-temporal, which is free to the shots order.

We argue that scene consistency is critical for the training on the unlabeled long-term videos due to the three reasons:
(i) the similar shots in the same scene may be far away;
(ii) the greater feature spacing between scenes is beneficial to the downstream task of \textit{Video Scene Segmentation}, and it can be achieved by maximizing inter-scene distance and minimizing intra-scene distance;
(iii) while the \textit{NN} selection may result in a trivial objective, due to the maximization of the similarity of the sample pairs that maybe already the closest, the scene consistency enables the selection to achieve a more non-trivial objective.

For the proposed scene consistency-based selection, we perform online clustering of samples in a batch, and use the cluster center sample as the positive sample with respect to the query shot, as shown in Fig. \ref{fig:positive} (d). The specified mapping function is formulated as follows: 
\begin{equation}
\begin{aligned}
 MAP_{SC}(i) = \arg \min _{j \in I_C} \left \| f(Q[i] \mid \theta_{Q})-f(Q[j] \mid \theta_{Q}) \right \|_2 
\end{aligned}
\label{eq:cluster}
\end{equation}
where $\ I_C = \{ic_{1}, ic_{2}, \ ... \ , ic_{\#class}\}$ stands for the indices of cluster centers, $\#class$ is the number of cluster centers. 

While center sample reflects the cluster-specified common information, we additionally use the query-specific individual information to generate the positive sample.
Unlike the conventional multiple-instance learning \cite{miech2020end}, which treats center and query samples as multiple positive samples, we propose to construct the soft positive sample, namely \textit{Soft-Scene Consistency (SC)} sample as follows:
\begin{equation}
\begin{aligned}
k_{Soft-SC} = \gamma k_{SA} + (1-\gamma) k_{SC}
\end{aligned}
\label{eq:soft}
\end{equation}
 where $\gamma$ is a trade-off parameter, $k_{SA}$ and $ k_{SC}$ are the key (positive) samples selected by \textit{Self-Augmented Selection} and \textit{Scene Consistency Selection}.

\textbf{Scene Consistency Data Augmentation.}
Since the early stage of training is not stable, too much color augmentations, e.g. grayscale transformations,
color jitter, etc., misguide the selection of positive samples, namely as \textit{Selection Shift}. In this case, the model focuses more on non-semantic information.
To solve this problem, some studies \cite{xu2021rethinking} directly omit color augmentations for better performance.
By contrast, we propose \textbf{Asymmetric Augmentation} to alleviate the influence of \textit{Selection Shift} and 
use color augmentation to further improve the performance.
More specifically, augmentations without the color transformation are used in $Aug_Q$ to get more accurate and scene consistent positive samples, while the color data augmentation operations are performed in $Aug_K$.

\textbf{Scene Agnostic Clip-Shuffling.} 
For fully leveraging the limited video data, we propose to construct more pseudo
scene cues. In this work, the data augmentation is based on the basic unit of clip, i.e. $\rho$ continuous shots, the generated clips are then randomly spliced disorderly for the training.

\begin{figure}[H]
  \centering
\includegraphics[width=.42\textwidth]{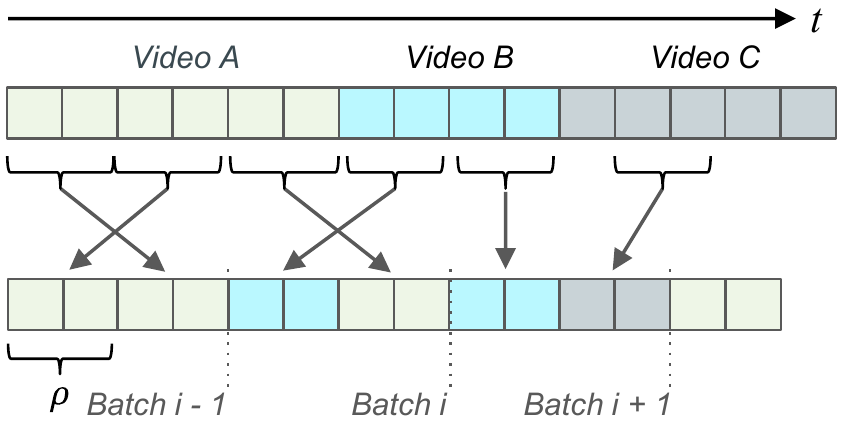}
  \caption{\textbf{The illustration of Scene Agnostic Clip-Shuffling.} Clips are spliced disorderly for training and each clip contains $\rho$ continuous shots.}
  \label{fig:Shuffling}
\end{figure}

The process of \textit{Scene Agnostic Clip Generation and Shuffling} is shown in Fig. \ref{fig:Shuffling}.

\subsubsection{Negative Sample Selections} 

The way to choose negative samples varies according to the specific SSL frameworks. For SimCLR \cite{chen2020simple}, the set of all non-positive samples within a batch is used as the negative samples, and MoCo \cite{chen2020improved} leverages a negative sample queue, which is a memory bank of previous samples output from the key encoder. 
However, BYOL \cite{grill2020bootstrap} and SimSiam \cite{chen2021exploring} do not use negative samples and instead resort to exploring more non-trivial solutions of SSL. 

\subsubsection{Objective Function} 

\textbf{With Negative Samples.} By defining $sim(\cdot,\cdot)$ as the cosine similarity, the contrastive loss
function, \textit{i.e.}, InfoNCE \cite{he2020momentum}  is employed and formulated as follows:
\begin{equation}
\begin{aligned}
{L_{con}}=-\log \frac{\sum_{k \in\left\{k^{+}\right\}} e^ {(\operatorname{sim}(q, k) / \tau)}}{\sum_{k \in\left\{k^{+}, k^{-}\right\}} e^ { (\operatorname{sim}(q, k) / \tau)}}
\end{aligned}
 \label{eq:nce}
\end{equation}
where ${k^{+}}$ and ${k^{-}}$ stand for the positive and negative samples for the query $q$, and the $\tau$ is the temperature term \cite{wu2018unsupervised}.

\textbf{Without Negative Samples.} By maximizing the similarity between the query and positive samples, the contrastive loss without negative samples is formulated as follows:
\begin{equation}
\begin{aligned}
{L_{con}}= -2\sum_{k \in\left\{k^{+}\right\}} (\operatorname{sim}(\mathcal{P}_\theta(q), k_{SG}) + \operatorname{sim}(\mathcal{P}_\theta(k), q_{SG}) )
\end{aligned}
  \label{eq:2}
\end{equation}
where $\mathcal{P}_\theta$ is the predictor $\mathcal{P}$ with parameters $\theta$  \cite{grill2020bootstrap,chen2021exploring}, $k_{SG}$ and $q_{SG}$ are the samples with stop-gradient (SG) \cite{grill2020bootstrap,chen2021exploring}.



\subsection{Video Scene Segmentation}
After the unsupervised pretraining,  two downstream models are used to evaluate the quality of the extracted features with the frozen query encoder.

\textbf{Problem Definition.} For the \textit{Video Scene Segmentation}, \cite{rao2020local} \cite{chen2021shot} convert the task into a binary classification task of shot semantics by modeling the temporal relationship of adjacent shot features. In this way, we can determine whether the end of a shot is the end of a scene story.

\begin{figure}[H]
  \centering
\includegraphics[width=.45\textwidth]{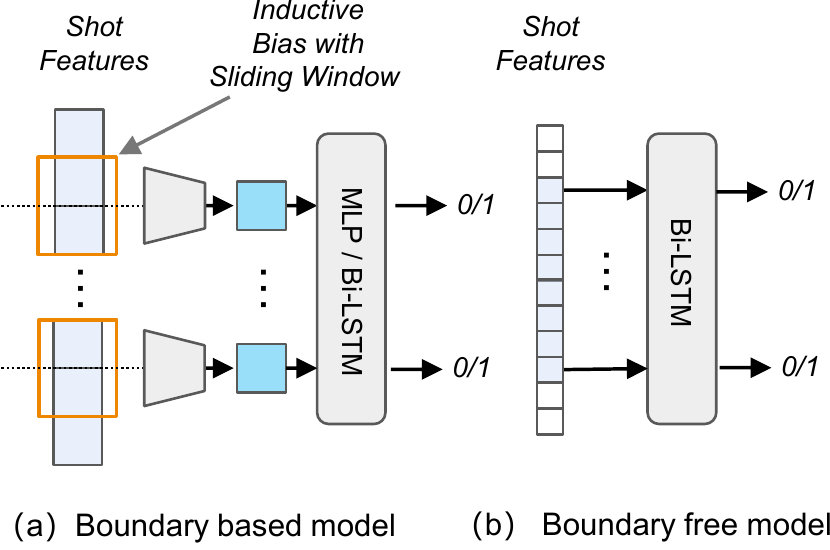}
  \caption{The illustration of boundary based model (a) and boundary free model (b) for \textit{Video Scene Segmentation}.}
  \label{fig:model}
\end{figure}

\textbf{Boundary free model.} While the previous downstream task of \textit{Video Scene Segmentation} is concluded to a shot boundary modeling based approach, as shown in Fig. \ref{fig:model} (a), we introduce a vanilla boundary-free model.
As shown in Fig. \ref{fig:model} (b), the proposed model covers the long-term dependence of shot representations based on sequence-to-sequence learning.
Compared with boundary based model in Fig. \ref{fig:model} (a) that introduces inductive bias for the shot boundary modeling with the sliding windows, the suggested model (b) takes the shot features as the basic temporal input unit, enabling the model to explore both local and global semantic relations.



\section{Experiments}

\subsection{Experimental Setup}

\textbf{Dataset.}
MovieNet \cite{huang2020movienet} consists of 1,100 movies with a large amount of multi-modal data and annotations, and the total duration of all movies is about 3000 hours, it is the 
largest dataset for movie understanding analysis by far. Besides, MovieNet \cite{huang2020movienet} is split into a training set with 660 movies, a validation set with 220 movies and a testing set with 220 movies. 
Currently, for the task of \textit{Video Scene Segmentation}, 190, 64 and 64 videos are labeled with scene boundaries for the training, validation and test sets, respectively. More importantly, \textit{\textbf{movies in the MovieScene \cite{rao2020local} are all included in MovieNet\cite{huang2020movienet}.}}

It is worth noting that there are two versions of annotation about \textit{Video Scene Segmentation} task associated with MovieNet, one with only 150 annotations (called \textit{MovieScenes} in \cite{rao2020local,chen2021shot}, used in earlier methods \cite{rao2020local} but it is no longer available), and one with a total of 318 annotations (abbreviated as \textit{MovieScenes-318} in this work). Since the small scale of of BBC \cite{baraldi2015deep} and OVSD \cite{rotman2017optimal} datasets and unavailability of AdCuepoints \cite{chen2021shot} dataset, we instead adopt MovieNet \cite{huang2020movienet} dataset to evaluate the related approaches, more details are in the \textit{Supplementary Materials}.

\begin{figure*}
\centering
\begin{minipage}{\textwidth}
\centering
\begin{minipage}[t]{0.40\textwidth}
\centering
\makeatletter\def\@captype{table}\makeatother\caption{Results of supervised methods w/o SSL for \\ the task of Video Scene Segmentation on MovieNet.}
{\scalebox{0.92}{
\begin{threeparttable}
\begin{tabular}{lccc}
\toprule[1.5pt]
\textbf{Methods} &  \textbf{Dataset} & \textbf{AP}  & \textbf{F1}\\
\toprule

SCSA \cite{chasanis2008scene}  & M.S.         & 14.7     & -     \\ 
Story Graph \cite{tapaswi2014storygraphs} &M.S.       & 25.1  & -     \\ 
Siamese \cite{baraldi2015deep} & M.S.        & 28.1   & -      \\ 
ImageNet \cite{deng2009imagenet}& M.S.       & 41.26   & -    \\ 
Places \cite{zhou2017places} & M.S.     & 43.23   & -     \\ 
LGSS \cite{rao2020local}   & M.S.          & 47.1   & -    \\ 
LGSS w/o DP \cite{rao2020local}    & M.S.     & 44.9  & -  \\ 
\midrule
LGSS w/o DP \cite{rao2020local} \tnote{*}    & M.S-318     & 44.9  & 38.52  \\ 
\bottomrule[1.5pt]
\end{tabular}
\begin{tablenotes}
    \footnotesize
    \item[*] Our implementations based on official public codebase,{\textcolor{red}{$^2$}} while \textit{DP (Dynamic Programming)} isn't public available.
  \end{tablenotes}
\end{threeparttable}
 \label{tab:sup}
}}
\end{minipage}
\begin{minipage}[t]{0.55\textwidth}
\centering
\makeatletter\def\@captype{table}\makeatother\caption{Results of methods w/ SSL for the task of Video Scene Segmentation on MovieNet.}
\centering
{\scalebox{0.89}{
\begin{threeparttable}
\begin{tabular}{lccccccc}
\toprule[1.5pt]
\textbf{Methods} & \textbf{Pretrain Data} & \textbf{Eval.} &\textbf{Protocol} & \textbf{AP}   &\textbf{F1} \\
\toprule

ShotCoL \cite{chen2021shot} & Train.+Test.+Val.  &  M.S   & MLP  \cite{chen2021shot}        & 52.83     & -       \\ 
ShotCoL \cite{chen2021shot} & Train.+Test.+Val.    &  M.S-318     & MLP \cite{chen2021shot}        & 53.37      & -        \\ 
ShotCoL \cite{chen2021shot} \tnote{*}  & Train.+Test.+Val.    &  M.S-318      & MLP \cite{chen2021shot}       & 52.89      & 49.17         \\ 
SCRL (ours)    & Train.+Test.+Val.    &  M.S-318       & MLP \cite{chen2021shot}      & \underline{54.82}    &   \underline{51.43}     \\ 
\midrule
ShotCoL \cite{chen2021shot} \tnote{*}   & Train. only &  M.S-318    & MLP \cite{chen2021shot}  & 46.77      &   45.78      \\ 
SCRL (ours)     & Train. only  &  M.S-318      & MLP \cite{chen2021shot}        & 53.74    &    50.40      \\ 
\midrule
ShotCoL \cite{chen2021shot} \tnote{*}   & Train. only &  M.S-318   & Bi-LSTM   & 48.21    & 46.52      \\ 
SCRL (ours)     & Train. only  &  M.S-318      & Bi-LSTM        & \textbf{54.55}    & \textbf{51.39}         \\ 
\bottomrule[1.5pt]
\end{tabular}
\begin{tablenotes}
    \footnotesize
    \item[*] Our implementations.
  \end{tablenotes}
\end{threeparttable}
\label{tab:unsup}
}}
\end{minipage}
\end{minipage}
\vspace{-1.0em}
\end{figure*}

\textbf{Representation Learning Stage.}
For visual modality, each shot consists of 3 keyframes and ResNet50 \cite{he2016deep} is chosen as the default backbone to learn the shot representations. The audio backbone used in \cite{rao2020local} is applied for audio modality, more details about the backbone encoders can be found in \textit{Supplementary Materials}.

For pretraining data, (i) training set (660 movies) in MovieNet \cite{huang2020movienet} is used to learn the shot representations, while we also conduct experiments with (ii) all data (1,100 movies) \cite{chen2021shot} for a fair comparison. In particular, although test data without the scene boundary labels are used for representation learning in setting (ii), \textbf{it is not recommended} to use all the data for pretraining because we usually have no prior access to test data in real scenarios. Moreover, for the most of Self-Supervised benchmarks \cite{he2020momentum,chen2020simple,chen2020big,chen2020improved,chen2021empirical,chen2021exploring,grill2020bootstrap,caron2020unsupervised}, representation learning is performed only on the training set, rather than all of the data.

\textbf{Video Scene Segmentation Stage.}
For existing Self-Supervised methods on images and videos, a simple downstream model is frequently used to evaluate the representation quality of the frozen encoders. For instance, a linear fully-connected layer is widely used for evaluation.
However, for the \textit{Video Scene Segmentation} task, we cannot determine whether the ending position of a single shot is the scene boundary or not. Consequently, a boundary-based non-temporal model (MLP-based protocol, followed by \cite{chen2021shot}) and a boundary-free temporal model (Bi-LSTM \cite{huang2015bidirectional}-based protocol, proposed by us) are employed to evaluate the capability of the encoder for local-to-global modeling.

\textbf{Metrics}. We use the mean of Average Precision (AP) \cite{rao2020local} \cite{chen2021shot} specified to ground truth scene boundaries of each movie, as well as F1-score for the evaluation.

\textbf{Implementation Details.}
During the  learning stage of Self-Supervised representation, the batch size is set to 1,024 (shots), initial learning rate is set to 0.03 and the training epoch is 100. The parameters of the visual and audio encoders are randomly initialized. Besides, we perform naive \textit{K-Means} algorithm \cite{lloyd1982least, liu2021group} for online clustering and the cluster number $\#class$ is set to 24, while the clip length, \textit{i.e.} $\rho$ of Scene Agnostic Clip Shuffling  is set to 16. 
MoCov2 \cite{chen2020improved} with the queue size of 65,536, momentum value of 0.999, temperature of 0.07 and cosine learning rate decay, are used as our SSL framework setting.
For the \textit{Video Scene Segmentation} task, $num$-$of$-$shot$ \cite{chen2021shot} is set to 4 and 40 for the MLP \cite{chen2021shot} and Bi-LSTM protocols, respectively. 
Each pretraining trial is conducted on the server with 8 NVIDIA V100 GPUs for approximate 24 hours in visual modality and 10 hours in audio modality.
The dimensions of visual and audio features used for both pretraining and evaluation are 2,048 and 512, respectively.
More details, \textit{e.g.}, the choice of hyperparameter, are presented in \textit{Supplementary Materials}.

\footnotetext{$^2$\url{https://github.com/AnyiRao/SceneSeg}}

\subsection{Comparison with Existing Methods}
Tables \ref{tab:sup} and \ref{tab:unsup} present an overall performance of methods w/ or w/o SSL for the \textit{Video Scene Segmentation} task, where \textit{M.S.} stands for \textit{MovieScenes} dataset with 150 annotated movies, and \textit{Eval.} means the dataset used for supervised \textit{evaluation stage} after the pretraining. Besides, \textit{Train., Test., and Val.} represent \textit{training, testing} and \textit{validation} sets of  MovieNet\cite{huang2020movienet}.

We have reproduced the performance of ShotCoL \cite{chen2021shot} on the entire dataset (1,100 movies) for comparison, although it is suggested to conduct the pretraining stage only on the training set. 
Compared with ShotCoL \cite{chen2021shot} that has a decline of 6.12 in terms of AP, our method can achieve competitive performance with less training data, with only a decline of 1.08 in terms of AP.
The proposed method outperforms the supervised state-of-the-art method, \textit{i.e.,} LGSS \cite{rao2020local} by margins of 9.65 in terms of AP and 12.87 in terms of F1.

\begin{figure*}
\centering
\begin{minipage}{\textwidth}
\centering
\begin{minipage}[t]{0.35\textwidth}
\centering
\makeatletter\def\@captype{table}\makeatother\caption{Ablation results of Positive Sample \\ Selection.}
{\scalebox{0.90}{
\begin{tabular}{l|cc}
\toprule[1.5pt]
\textbf{Methods} &  \textbf{Selection Strategy}   &  \textbf{AP} \\
\toprule
MoCo \cite{chen2020improved} & Self-Augmented       & 42.51           \\ 
- & Random ($n=1$)  & 43.24             \\ 
ShotCol \cite{chen2021shot} & NN ($m=8$)   & 46.77          \\ 
\midrule
SC & Scene Consistency    & \textbf{49.71}             \\ 
\bottomrule[1.5pt]
\end{tabular}
 \label{tab:ab_pos_select}
}}
\end{minipage}
\begin{minipage}[t]{0.32\textwidth}
\centering
\makeatletter\def\@captype{table}\makeatother\caption{Ablation results of SSL methods \\  w/ and w/o Scene Agnostic Clip-Shuffling.}
\centering
{\scalebox{0.9}{
\begin{tabular}{l|c|c|c}
\toprule[1.5pt]
\textbf{Methods} &  w/o  & w/  & \textbf{AP} \\
\toprule
NN  & \checkmark  & $\times$  &    46.77  \\ 
NN & $\times$  & \checkmark   &   48.63 \\
\midrule
SC &\checkmark & $\times$   &  49.71        \\ 
SC & $\times$   & \checkmark  & \textbf{52.17}   \\ 
\bottomrule[1.5pt]
 \end{tabular}
 \label{tab:shuffling}
}}
\end{minipage}
\begin{minipage}[t]{0.3\textwidth}
\centering
\makeatletter\def\@captype{table}\makeatother\caption{Ablation results of Multiple Positive Samples (MPS).}
\centering
{\scalebox{0.84}{
\begin{tabular}{l|c|c}
\toprule[1.5pt]
\textbf{Methods} &  \textbf{Positive Sample(s)}  &  \textbf{AP}   \\
\toprule
SC  & Center   &  52.17       \\ 
MPS-SC  & Self and Center    &   51.20        \\ 
Soft-SC & Eq. \ref{eq:soft} ($\gamma$=0.5)    &  \textbf{53.74}           \\ 

\bottomrule[1.5pt]
 \end{tabular}
 \label{tab:MIL}
}}
\end{minipage}
\end{minipage}
\end{figure*}

\subsection{Ablation Study}
We perform all the ablation experiments using only the training data of MovieNet in SSL stage, and evaluate the performance on downstream task based on MLP protocol for fairness.

\begin{figure}[H]
  \centering
   \includegraphics[width=0.98\linewidth]{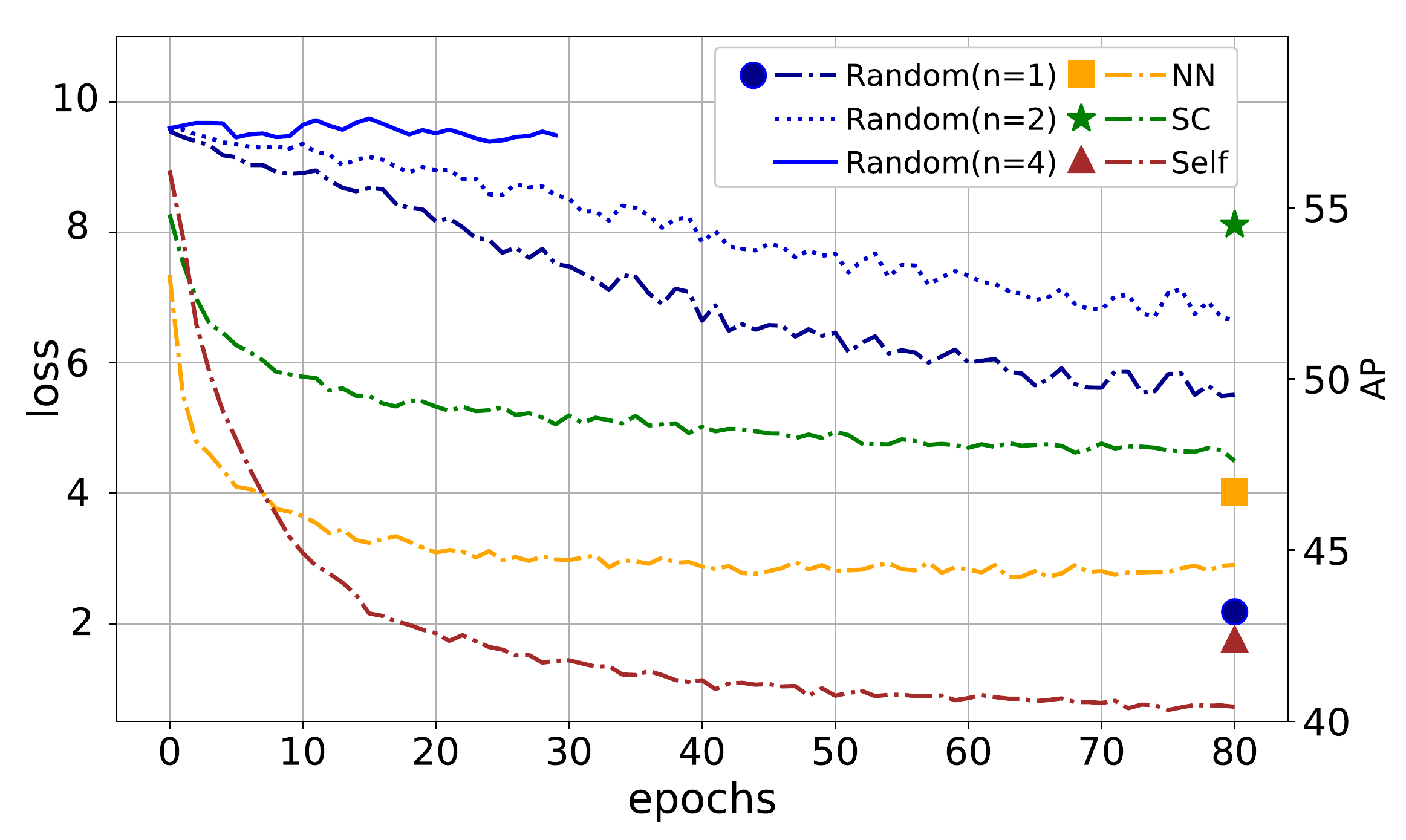}
   \caption{\textbf{Loss evolution curves and AP results} of the training with different selection strategies.}
   \label{fig:pretrain_curve}
\end{figure}

\textbf{Positive Sample Selection.} We first conduct ablation experiments on the four different selection methods of positive pairs, \textit{i.e.}, \textit{Self-Augmented Selection}, \textit{Random Selection}, \textit{Nearest Neighborhood (NN) Selection} and \textit{Scene Consistency (SC) Selection}. 
Tab. \ref{tab:ab_pos_select} shows that \textit{Scene Consistency Selection} method achieves better performance than the other selection methods, which outperforms the state-of-the-art algorithm \cite{chen2021shot} by a margin of 2.95 in terms of AP.
Meanwhile, the loss evolution curves of above methods are shown in Fig. \ref{fig:pretrain_curve}.
We can find that \textit{Self-Augmented Selection} reaches the lowest loss value, while obtaining the worst performance on the task of \textit{Video Scene Segmentation}.
Due to the trivial objective introduced by \textit{NN Selection} that is discussed in Section \ref{sec:SC}, it achieves the fastest convergence rate during the early training, while stagnating to a mediocre performance.
By contrast, \textit{SC Selection} has a relatively moderate convergence rate, and achieves the best performance among all the selection strategies.

\textbf{Scene Agnostic Clip-Shuffling.} 
The results of ablation study specific to the Scene Agnostic Clip-Shuffling are presented in Tab. \ref{tab:shuffling}.
Tab. \ref{tab:shuffling} shows that the  proposed \textit{Clip-Shuffling} achieves improvements of 1.85 and 2.46 in terms of AP for \textit{NN} and \textit{SC} methods, respectively. These results verify the advantage of the proposed positive sample selection discussed in Section \ref{sec:SC} that \textit{SC} is free to the shot order in a video.  

\textbf{Multiple Positive Samples (MPS).} Moreover, we study the performance of multiple positive samples in Tab. \ref{tab:MIL}. As shown in Tab. \ref{tab:MIL}, \textit{Soft-SC} achieves the best performance of 53.74 in terms of AP. Although single positive sample is employed in SC, it still achieves better performance than MPS-SC that employ multiple positive samples.

\begin{figure*}
  \centering
   \includegraphics[width=0.70\linewidth]{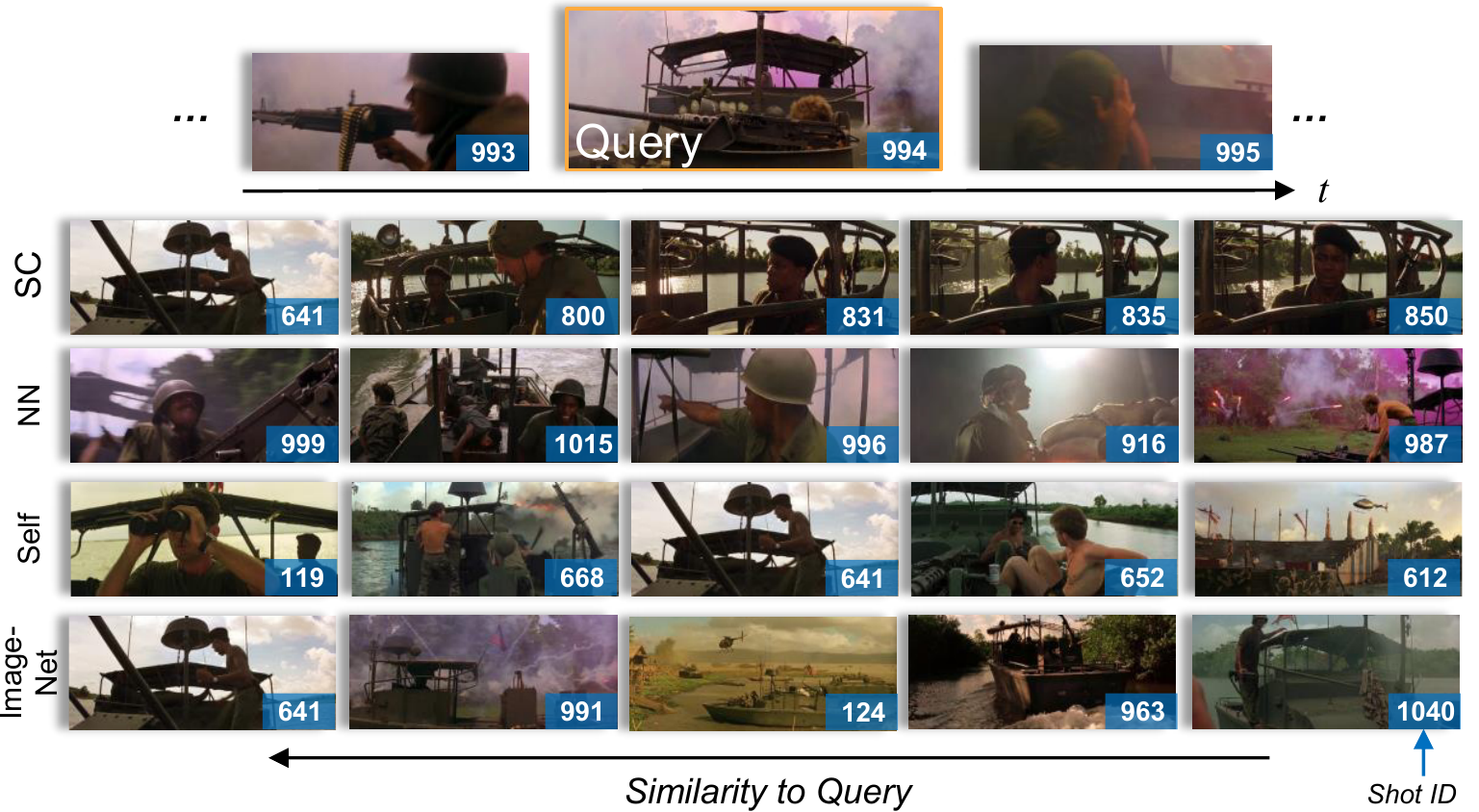}

   \caption{\textbf{The visualization results of shot retrieval}. 
   Overall,
   \textbf{\textit{NN}} tends to select adjacent shots, \textbf{\textit{Self}} shows less relevance to the query and \textbf{\textit{ImageNet}} retrieves many kinds of boats.
   Compared with the other methods, the results of \textbf{\textit{SC}} are more consistent in the semantic information, \textit{i.e.}, there is a man staying in the boat, and \textbf{\textit{SC}} achieves a larger span (\textit{i.e.}, from $641$ to $850$) than \textbf{\textit{NN}} according to shot IDs.
   Meanwhile, \textbf{\textit{SC}} shows better robustness against the interference of pink smoke in the $994$-th shot as the results are more pure.
   }
   \label{fig:retrieval}
\end{figure*}

\subsection{Analysis of the Proposed Method}

\textbf{Generalizability to the large-scale supervised approach.} To study the generalizability of the proposed method, we equip our trained models with LGSS \cite{rao2020local}, where LGSS is a large-scale supervised method and utilizes various pretrained models with multi-modality. As is Shown in 
Tab. \ref{tab:Generalizability}, our trained model, trained only on the unlabeled training set, and based on the same backbone, \textit{i.e.} ResNet-50 \cite{he2016deep}, achieves an improvement of 4.0 in terms of AP over the approach without our trained model.

\begin{table}[H]
\centering
\caption{Generalizability to the large-scale supervised approach. }
{\scalebox{0.9}{

\begin{tabular}{l|c|c}
\toprule[1.5pt]
\textbf{Methods} &\textbf{Modalities} & \textbf{AP} \\
\midrule
LGSS \cite{rao2020local} w/o SSL  & \multicolumn{1}{c|}{\begin{tabular}[c]{@{}c@{}} \textbf{Visual(Place, ResNet50)} \\ +Action+Actor+Audio \end{tabular}} & 44.9  \\ 
\midrule
LGSS \cite{rao2020local} w/ SSL  &  \multicolumn{1}{c|}{\begin{tabular}[c]{@{}c@{}} \textbf{Visual(SSL, ResNet50)} \\ +Action+Actor+Audio \end{tabular}} & \textbf{48.9} \\ 
\bottomrule[1.5pt]
\end{tabular}
}}
\label{tab:Generalizability}
\end{table}

\textbf{Performance on different Self-Supervised Learning (SSL) frameworks.} 
Four popular SSL frameworks are used for evaluating our method, \textit{i.e.}, SimCLR \cite{chen2020simple}, MoCo \cite{chen2020improved}, BYOL \cite{grill2020bootstrap} and SimSiam \cite{chen2021exploring}. 
Tab. \ref{tab:SSL} shows that the SSL framework with momentum updates and
negative samples achieves the best performance for the \textit{Video Scene Segmentation} task.
Due to the momentum update mechanism, the proposed method embedded in the framework of BYOL \cite{grill2020bootstrap} achieves an improvement of 10.53 over that in SimSiam \cite{chen2021exploring}, and a similar conclusion is reached in \cite{feichtenhofer2021large}.

\begin{table}[H]
\centering
\caption{Results of the proposed method based on various Self-Supervised Frameworks.}
{\scalebox{0.80}{
\begin{tabular}{l|c|c|c|c}
\toprule[1.5pt]
Methods  & \multicolumn{1}{c|}{\begin{tabular}[c]{@{}c@{}} SSL Fra-\\ meworks  \end{tabular}} &
\multicolumn{1}{c|}{\begin{tabular}[c]{@{}c@{}}w/ negative \\ samples  \end{tabular}} & 
\multicolumn{1}{c|}{\begin{tabular}[c]{@{}c@{}}w/ momen-\\tum update \end{tabular}} & 
 \textbf{AP}    \\ 
\midrule
SCRL  & SimSiam \cite{chen2021exploring} &       $\times$        &        $\times$       & 39.82 \\
SCRL  & SimCLR \cite{chen2020simple}  &       \checkmark          &     $\times$          & 45.32 \\ 
SCRL  & BYOL \cite{grill2020bootstrap}   &        $\times$          &         \checkmark         & 50.35 \\ 
\midrule
ShotCoL \cite{chen2021shot} & MoCo \cite{chen2020improved}   &        \checkmark       &      \checkmark       & 46.77 \\ 
SCRL & MoCo \cite{chen2020improved}  &        \checkmark       &      \checkmark       & \textbf{53.74} \\ 
\bottomrule[1.5pt]

\end{tabular}
\label{tab:SSL}
}}
\end{table}

\textbf{Boundary free model for evaluation.} 
To study the performance of the introduced boundary free model, the proposed method under MLP and Bi-LSTM protocols for the scene segmentation task is evaluated in Tab. \ref{tab:downstream}.
Since Bi-LSTM protocol has less inductive bias than \textit{sliding window} based MLP protocol, it is able to model representations of longer shot, hence achieves better performance on the task of \textit{Video Scene Segmentation}. More specifically, the performance of Bi-LSTM protocol increases as the length of the shots increases, while the performance of MLP protocol decreases instead. More details can be found in \textit{Supplementary Materials}.

\begin{table}[H]
\centering
\caption{Results of \textit{Video Scene Segmentation} using the proposed method under MLP and Bi-LSTM protocols.}
{\scalebox{0.82}{

\begin{tabular}{l|c|lll}
\toprule[1.5pt]
\textbf{Protocols} &\textbf{Shot-Len} & \textbf{AP} & \textbf{F1}  &  \textbf{\#Param} \\
\midrule
MLP \cite{chen2021shot} & 4 & 53.74 & 50.40 & 37.75 M  \\ 
MLP \cite{chen2021shot} & 10 & 49.61 \textcolor{red}{$\downarrow$} & 44.04 \textcolor{red}{$\downarrow$}  & 88.09 M  \textcolor{red}{$\uparrow$}  \\
\midrule
Bi-LSTM & 10 & 43.94 &  42.12 & 15.22 M  \\ 
Bi-LSTM & 40 & \textbf{54.55} \textcolor{red}{$\uparrow$} &  \textbf{51.39} \textcolor{red}{$\uparrow$} & 15.22 M \\ 
\bottomrule[1.5pt]
\end{tabular}
}}
\label{tab:downstream}
\end{table}

\textbf{Visualization of Shot Retrieval.} To get more intuition for the proposed selection, we conduct retrieval experiments using four selection methods, \textit{i.e.}, \textit{SC, NN, Self-Augmented and ImageNet selections}, and present the results in Fig. \ref{fig:retrieval}. 
More specifically, for a given shot, we calculate the similarities between it and the other shots in the entire movie, then visualize the TOP-5 most similar shots in Fig. \ref{fig:retrieval}.

\subsection{Limitations}

\textbf{Multi-modal Pretraining.} In order to test the performance of the proposed algorithm generalizing to multi-modal data, we also conduct experiments with audio and visual modalities in the SSL stage, the joint multi-modal learning scheme follows \cite{alayrac2020self}. However, 
we did not achieve any improvement and were confronted with the same concern that is mentioned in \cite{chen2021shot}, as shown in Table. \ref{tab:multi-modal}. Possible reasons are that (i) the publicly available audio data of each shot are incomplete, (ii) the raw audio data are not available yet due to copyright restrictions \cite{huang2020movienet} and (iii) LGSS \cite{rao2020local} utilizes various pretrained models on the other datasets, while the methods in the comparison are trained from scratch. 
Therefore, it is meaningful to shed light on how to pretrain better multi-modal representations on the MovieNet \cite{huang2020movienet}.

\begin{table}[H]
\centering
\caption{AP results of the multi-modal experiment on MovieNet. Backbones of following methods for each modality are the same.}
{\scalebox{0.86}{
\begin{tabular}{l|ccc}
\toprule[1.5pt]
\textbf{Methods}     & \textbf{Visual} & \textbf{Audio}   & \textbf{Visual+Audio}\\
\midrule
LGSS \cite{rao2020local} & 39.0          &  17.5  &  43.4    \\ 
\midrule
ShotCoL \cite{chen2021shot}        & 46.77          &  27.92  &  44.32    \\ 
SCRL         & 53.74          &  29.39  & 50.80      \\ 
\bottomrule[1.5pt]
\end{tabular}
}}
\label{tab:multi-modal}
\end{table}

\section{Conclusion}

We present a Self-Supervised Learning (SSL) scheme based on Scene Consistency to obtain better shot representations for the unlabeled long-term videos. The proposed method achieves the state-of-the-art performance on the task of \textit{Video Scene Segmentation} under various protocols, and significant better generalization performance when it is equipped with large-scale supervised approach. Besides, we introduce a fair pretraining protocol and a more comprehensive evaluation metric for the task of \textit{Video Scene Segmentation}, to make the assessment of the SSL more meaningful in practice.
~\\

\noindent \textbf{Acknowledgment}
The work was supported by the National Natural Science Foundation of China under grants no. 61602315, 91959108, the Science and Technology Project of Guangdong Province under grant no. 2020A1515010707,  the Science and Technology Innovation Commission of Shenzhen under grant no. JCYJ20190808165203670.

{\small
\bibliographystyle{ieee_fullname}
\bibliography{main}
}

\end{document}


\title{Scene Consistency Representation Learning for Video Scene Segmentation \\ 
Supplementary Materials}

\author{Haoqian Wu$^{1,2,3,4*}$, Keyu Chen$^{2*}$, Yanan Luo$^{2}$, Ruizhi Qiao$^{2}$, Bo Ren$^{2}$, \\ 
Haozhe Liu$^{1,3,4,5}$, Weicheng Xie$^{1,3,4}{}^\dagger$, Linlin Shen$^{1,3,4}$ \\
$^1$ Computer Vision Institute, Shenzhen University $^2$ Tencent YouTu Lab \\
$^3$ Shenzhen Institute of Artificial Intelligence and Robotics for Society \\
$^4$ Guangdong Key Laboratory of Intelligent Information Processing $^5$ KAUST \\
{\tt\small wuhaoqian2019@email.szu.edu.cn}
{\tt\small \{yolochen, ruizhiqiao, timren\}@tencent.com} \\ 
{\tt\small luoyanan93@gmail.com} 
{\tt\small haozhe.liu@kaust.edu.sa}
{\tt\small \{wcxie, llshen\}@szu.edu.cn}
}


\maketitle

\section{Additional Illustrations}

\subsection{Frame, Shot, Clip and Scene}
\begin{figure}[H]
  \centering
\includegraphics[width=.35\textwidth]{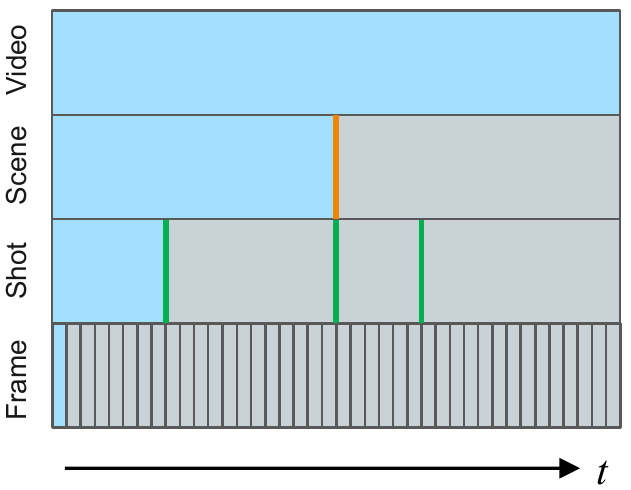}
  \caption{\textbf{The illustration of the connection between frame, shot, scene and video.} The \textcolor[RGB]{0,176,80}{green} and \textcolor[RGB]{239,134,0}{orange} lines represent \textit{Shot} and \textit{Scene Boundaries}, respectively.}
  \label{fig:shots}
\end{figure}

Fig. \ref{fig:shots} shows the connection between \textbf{frame}, \textbf{shot}, \textbf{scene} and \textbf{video}. More specifically, a \textbf{shot} contains only continuous frames taken by the camera without interruption, and a \textbf{scene} is composed of successive shots and describes a same short story. Typically, a consecutive shot sequence of arbitrary length can be treated as a \textbf{clip}.

\footnotetext{\noindent$^*$Equal Contribution \quad $^\dagger$Corresponding Author}

\subsection{Shot Sampling Strategy}

Video data is highly redundant because there are many repetitive frames in chronological order, and we follow the sampling strategy in \cite{rao2020local,chen2021shot} for long-term videos.
More concretely, the video sequence is sliced according to the shot boundaries that are determined by the transition of the visual modality. Based on the beginning and ending positions of shots, a fixed number of $N=3$ frames are selected and treated as the original feature of one shot, where the starting, middle and ending frames of the shot are sampled.

\section{Algorithm Details}

\subsection{Representation Learning Stage}
In the Representation Learning Stage, the proposed SSL algorithm can be summarized in Algorithm \textcolor{red}{1}.

\begin{algorithm}[!htb]
    \caption{SSL for Video Scene Segmentation.}
    \textbf{Input:} Training samples $X$ \\
    \textbf{Require:} Initialized encoders $f\left(\cdot \mid \theta_{Q}\right)$ and $f\left(\cdot \mid \theta_{K}\right)$; Initialized memory bank $Queue$; Augmentation operations $Aug_Q(\cdot)$ and $Aug_{K}(\cdot)$; Number of iterations $n_{iter}$
  \begin{algorithmic}[1]
    \For {$i=1$ to $n_{iter}$}
      \State Obtain augmented training samples $Q, K$ by $Q=Aug_Q(X), K=Aug_{K}(X)$;
      \State Obtain mapping function $MAP(\cdot)$;
      \State Obtain positive pairs $\{q, k^{+}\}$ by Eq. \textcolor{red}{(1)};
      \State Detach samples $k^{+}$  from Calculation Graph;
      \State Obtain negative samples $k^{-}$ from $Queue$;
      \State Calculate contrastive loss $L_{con}$  by Eq. \textcolor{red}{(7)} or \textcolor{red}{(8)};
      \State Perform backpropagations for $L_{con}$;
      \State Update encoder $f\left(\cdot \mid \theta_{Q}\right)$ by gradient descent;
      \State Update encoder $f\left(\cdot \mid \theta_{K}\right)$ by momentum update;
      \State Enqueue the positive samples $k^{+}$ to $Queue$;
    \EndFor
\end{algorithmic}
    \textbf{Output:} Query encoder $f\left(\cdot \mid \theta_{Q}\right)$.
\label{algo:1}
\end{algorithm}

\subsection{Video Scene Segmentation Stage}
\begin{figure}[H]
  \centering
\includegraphics[width=.45\textwidth]{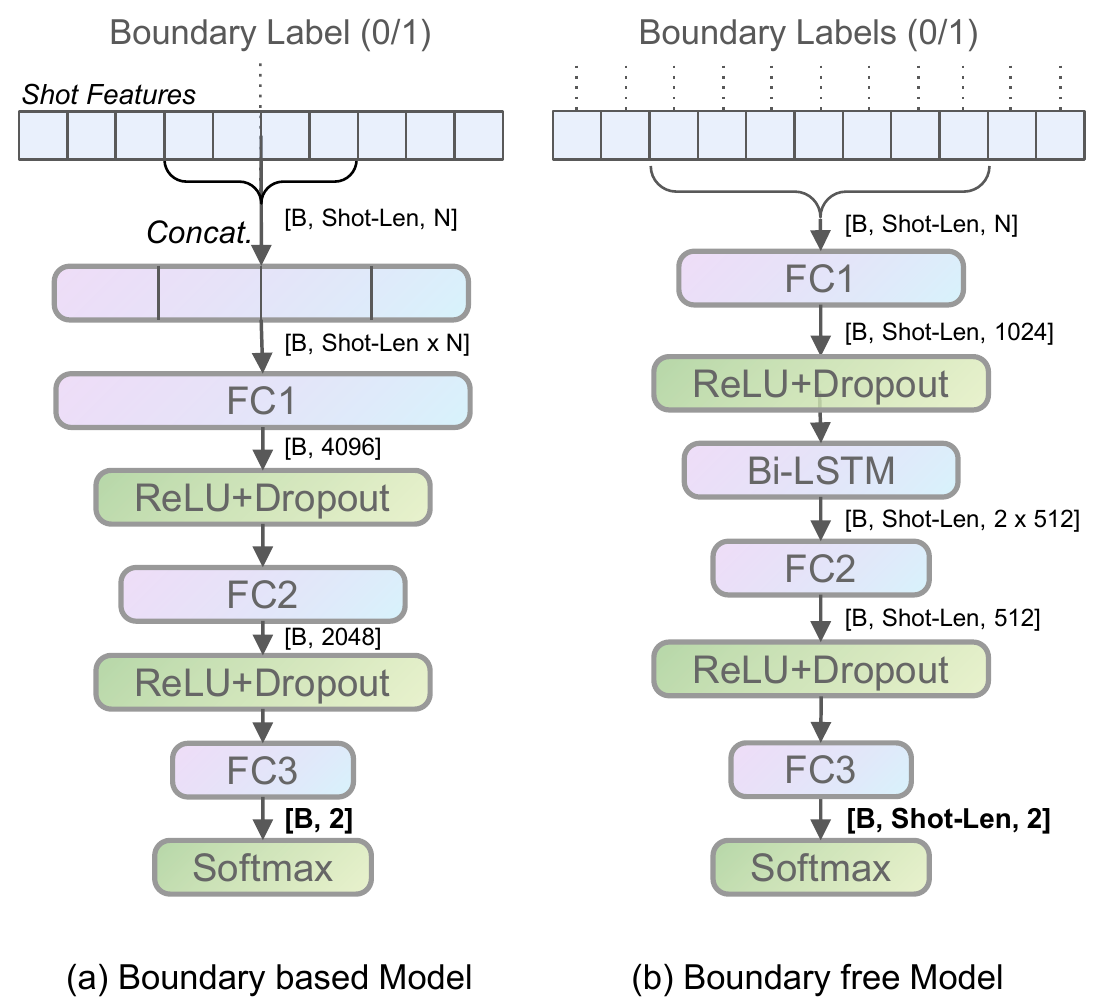}
  \caption{The illustration of Boundary based and Boundary free models, where $B$, $N$ and $Shot$-$Len$  represent the batch size, dimension of the feature and length of shots processed within a batch.}
  \label{fig:models}
\end{figure}

The Boundary based model (\textit{i.e.}, MLP protocol \cite{chen2021shot}) and Boundary free model (\textit{i.e.}, Bi-LSTM protocol introduced by us) are presented in Fig. \ref{fig:models}.

For the  MLP protocol \cite{chen2021shot}, we use SGD as the optimizer, and the weight decay is 1e-4 and the SGD momentum is 0.9. In the training stage, we use a mini-batch size of 128 and dropout rate of 0.4 for FC layers. Besides, we train for 200 epochs with the learning rate multiplied by 0.1 at 50, 100 and 150 epochs.

For the  Bi-LSTM protocol, we use SGD as the optimizer, and the weight decay is 1e-4 and the SGD momentum is 0.9. In the training stage, we use a mini-batch size of 32 and dropout rate of 0.7 for FC layers (except for the last layer), the Bi-LSTM was implemented using the LSTM module in PyTorch \cite{paszke2019pytorch}, which includes two layers with 512 hidden units together with a dropout layer with the dropout probability of 0.6 between them. Besides, we train for 200 epochs with the learning rate multiplied by 0.1 at 160, 180 epochs.
In the inference stage, in order to make each shot aggregate as much information as possible from the adjacent shots, in each inference batch, we use the middle portion of the model output sequence as the scene boundary. More specifically, for the input shot feature sequence, \textit{i.e.} $[S_0,S_1, \cdots, S_{Shot-Len\ -1}]$, we adopt the subsequence, \textit{i.e.} $[Y_{\lceil Shot-Len/4 \rceil}, \cdots, Y_{\lceil 3*Shot-Len/4 \rceil \ -1}]$ of the corresponding output sequence, \textit{i.e.} $[Y_0,Y_1,  \cdots, Y_{Shot-Len\ -1}]$ as the prediction result. Additionally, the first and last shot features are used to pad the beginning and ending of the shot feature sequence, respectively.

\section{Additional Implementation Details}

\subsection{Datasets}
The details of BBC \cite{baraldi2015deep}, OVSD \cite{rotman2017optimal}, MovieNet \cite{huang2020movienet} and MovieScene \cite{rao2020local} datasets are shown in Tab. \ref{tab:datasets}. The scalability of BBC \cite{baraldi2015deep} and OVSD \cite{rotman2017optimal} is much smaller than that of MovieNet \cite{huang2020movienet} and MovieScene \cite{huang2020movienet}. BBC \cite{baraldi2015deep} has 5 different annotations and the number of scenes is averaged, as shown in Tab. \ref{tab:datasets}. The ground truth Shot Detection result of OVSD \cite{rotman2017optimal} is unavailable from its official website and \cite{rao2020local}, thus, we provide our implementation results on Shot Detection in the Tab.  \ref{tab:datasets}.
It is worth noting that \textit{\textbf{movies in the MovieScene \cite{rao2020local} are all included in MovieNet\cite{huang2020movienet}.}}

\begin{table}[H]
\centering
\caption{Details of BBC/OVSD/MovieNet/MovieScene datasets.}
\scalebox{0.8}{

\begin{tabular}{l|r|rrr}
\toprule[1.5pt]
\textbf{Datasets} & \textbf{\#Video}  &\textbf{Time(h)} & \textbf{\#Shot} & \textbf{\#Scene} \\
\midrule
BBC \cite{baraldi2015deep}  & 11 & 9 & 4,900 & 547 \\ 
OVSD \cite{rotman2017optimal} & 21 & 10 & 9,377 & 607\\ 
MovieScene-150 \cite{rao2020local} & 150 & 297 & 270,450 & 21,428 \\ 
MovieScene-318 \cite{rao2020local} & 318 & 601 & 503,522 & 41,963 \\ 
\midrule
MovieNet \cite{huang2020movienet}  & 1,100 & 3,000 &  &  \\ 

\bottomrule[1.5pt]
\end{tabular}
}
\label{tab:datasets}
\end{table}

To show the difference of the 5 annotation results, we present the average (mean), minimum (min), maximum (max) and standard deviation (std.) of the number of scenes for each video in BBC \cite{baraldi2015deep} with each annotation in Tab. \ref{tab:BBC_all}.

\begin{table}[H]
\centering
\caption{Number of scenes in each video of BBC dataset.}
\scalebox{0.9}{

\begin{tabular}{c|cccc}
\toprule[1.5pt]
\multirow{2}{*}{\textbf{Video Names}} & \multicolumn{4}{c}{\#Scene} \\ \cline{2-5} 
 & \textbf{mean}  &\textbf{min} & \textbf{max} & \textbf{std.}  \\
\midrule
From Pole to Pole  & 47.8 & 23 & 65 & 14.4 \\ 
Mountains & 47.6 & 36 & 62 & 9.8\\ 
Ice Worlds & 50.6 & 33 & 69 & 12.1 \\  
Great Plains & 52.0 & 30 & 74 & 14.1 \\ 
Jungles & 47.4 & 25 & 59 & 11.8 \\ 
Seasonal Forests & 53.4 & 33 & 71 & 12.1 \\ 
Fresh Water  & 55.2 & 37 & 70 & 10.5 \\
Ocean Deep  & 47.8 & 29 & 67 & 12.7 \\ 
Shallow Seas & 49.2 & 33 & 66 & 12.0 \\ 
Caves & 45.4 & 22 & 63 & 14.1 \\ 
Deserts & 50.8 & 26 & 64 & 13.7 \\ 
\midrule
\textbf{All Videos Avg.}  & 547 &  &  &  \\ 
\bottomrule[1.5pt]
\end{tabular}
}
\label{tab:BBC_all}
\end{table}

\subsection{Backbones}
For the visual modality, ResNet50 \cite{he2016deep} is used as the encoder with the same modification \cite{chen2021shot} that the input channel number of first convolution layer is changed from 3 to 9. As for audio modality, we adopt the same backbone used in \cite{rao2020local}.

\subsection{Choice of Hyperparameters}

Two hyperparameters are introduced in the Sec. \textcolor{red}{3.1.2} of the main body of this work, \textit{i.e.} number of cluster ($\#class$) for Scene Consistency Selection strategy and length of continuous shots ($\rho$) for Scene Agnostic Clip-Shuffling. 
We study the sensitivity of the proposed algorithm against these two hyperparameters
in Tab. \ref{tab:hyperparameters}. MLP \cite{chen2021shot} protocol on the MovieScene-318 dataset is used in this experiment.

\begin{table}[H]
\centering
\caption{AP results for different settings of hyperparameters. The \textbf{bolded} and \underline{underlined} values stand for the optimal and suboptimal performances, respectively.}
\scalebox{0.9}{

\begin{tabular}{l|ccc}
\toprule[1.5pt]
\textbf{$\rho\ $/$\#class$} & \textbf{16}  &\textbf{24} & \textbf{32}  \\
\midrule
\textbf{16}  & 51.80 & \textbf{53.74}  & 53.22 \\ 
\textbf{24}  & 52.64 & \underline{53.62} & 53.13\\ 
\textbf{32}  & 52.68 & 52.91 & 53.05\\ 


\bottomrule[1.5pt]
\end{tabular}
}
\label{tab:hyperparameters}
\end{table}

\subsection{Data Augmentation Details}

We follow the  data augmentation operation used in the \cite{chen2020simple}, \textit{i.e.} random cropping, flipping, color distortion, Gaussian bluring. A PyTorch-like pseudo code for the data augmentations, \textit{i.e., Asymmetric Augmentation mentioned in  Sec. \textcolor{red}{3.1.2} of the body of this work}, is presented as follows:
\begin{lstlisting}[language=Python, caption=A PyTorch-like pseudo code for the data augmentation.]
import torchvision.transforms as transforms
normalize = transforms.Normalize(
    mean=[0.485, 0.456, 0.406],
    std=[0.229, 0.224, 0.225]) 
# augmentation for key encoder
augmentation_key_encoder = [
    transforms.ToPILImage(),
    transforms.RandomResizedCrop(224, scale=(0.2, 1.)),
    transforms.RandomApply([
        transforms.ColorJitter(0.4, 0.4, 0.4, 0.1)], p=0.5),
    transforms.RandomGrayscale(p=0.2),
    transforms.RandomApply([GaussianBlur([.1, 2.])], p=0.5),
    transforms.RandomHorizontalFlip(),
    transforms.ToTensor(),
    normalize
    ]
# augmentation for query encoder
augmentation_query_encoder = [
    transforms.ToPILImage(),
    transforms.RandomResizedCrop(224, scale=(0.2, 1.)),
    transforms.RandomApply([GaussianBlur([.1, 2.])], p=0.5), 
    transforms.RandomHorizontalFlip(),
    transforms.ToTensor(),
    normalize
    ]
\end{lstlisting}

\section{Additional Results}

\subsection{Results on BBC/OVSD Datasets}
\label{sec:bbc}

 Since the training/validation/testing datasets of BBC \cite{baraldi2015deep}/OVSD \cite{rotman2017optimal} are not available and the scale of these two datasets is very small compared to MovieNet \cite{huang2020movienet} dataset, we apply the model trained on MovieNet \cite{huang2020movienet} onto BBC \cite{baraldi2015deep} and OVSD \cite{rotman2017optimal} to study the generalization abilities of the algorithms without the finetuning, the results are shown in Tab. \ref{tab:OVSD} and Tab. \ref{tab:BBC}.

Tab. \ref{tab:OVSD} shows that the proposed method outperforms ShotCoL \cite{chen2021shot} by a large margin of 13.27 in terms of AP on OVSD \cite{rotman2017optimal}. 

\begin{table}[H]
\centering
\caption{AP results on OVSD dataset. }
\scalebox{0.9}{

\begin{tabular}{l|c}
\toprule[1.5pt]
\textbf{Methods} & \textbf{AP}   \\
\midrule
ShotCoL \cite{chen2021shot}  & 25.53  \\ 
SCRL  & \textbf{38.80} \\ 

\bottomrule[1.5pt]
\end{tabular}
}
\label{tab:OVSD}
\end{table}

We conduct experiments of 5 different annotators on BBC \cite{baraldi2015deep} and show the average performances in Tab. \ref{tab:BBC}, where the proposed method outperforms the compared method by a margin of 2.20 in terms of AP.

\begin{table}[H]
\centering
\caption{AP results on BBC dataset. \textbf{A. \textit{i}} stands for the $i$-th annotation and \textbf{Avg.} represents the average of the results of 5 different annotators.}
\scalebox{0.8}{

\begin{tabular}{l|ccccc|c}
\toprule[1.5pt]
\textbf{Methods} & \textbf{A. 1}  & \textbf{A. 2} & \textbf{A. 3} & \textbf{A. 4} & \textbf{A. 5} & \textbf{Avg.}  \\
\midrule
ShotCoL \cite{chen2021shot}  & 29.90 & 30.81 & 31.45 & 26.45 & 21.27 & 27.98 \\ 
SCRL  & 32.45 & 32.54 & 33.27 & 28.36 & 24.27 & \textbf{30.18}\\ 

\bottomrule[1.5pt]
\end{tabular}
}
\label{tab:BBC}
\end{table}

\begin{figure}[H]
  \centering
\includegraphics[width=.45\textwidth]{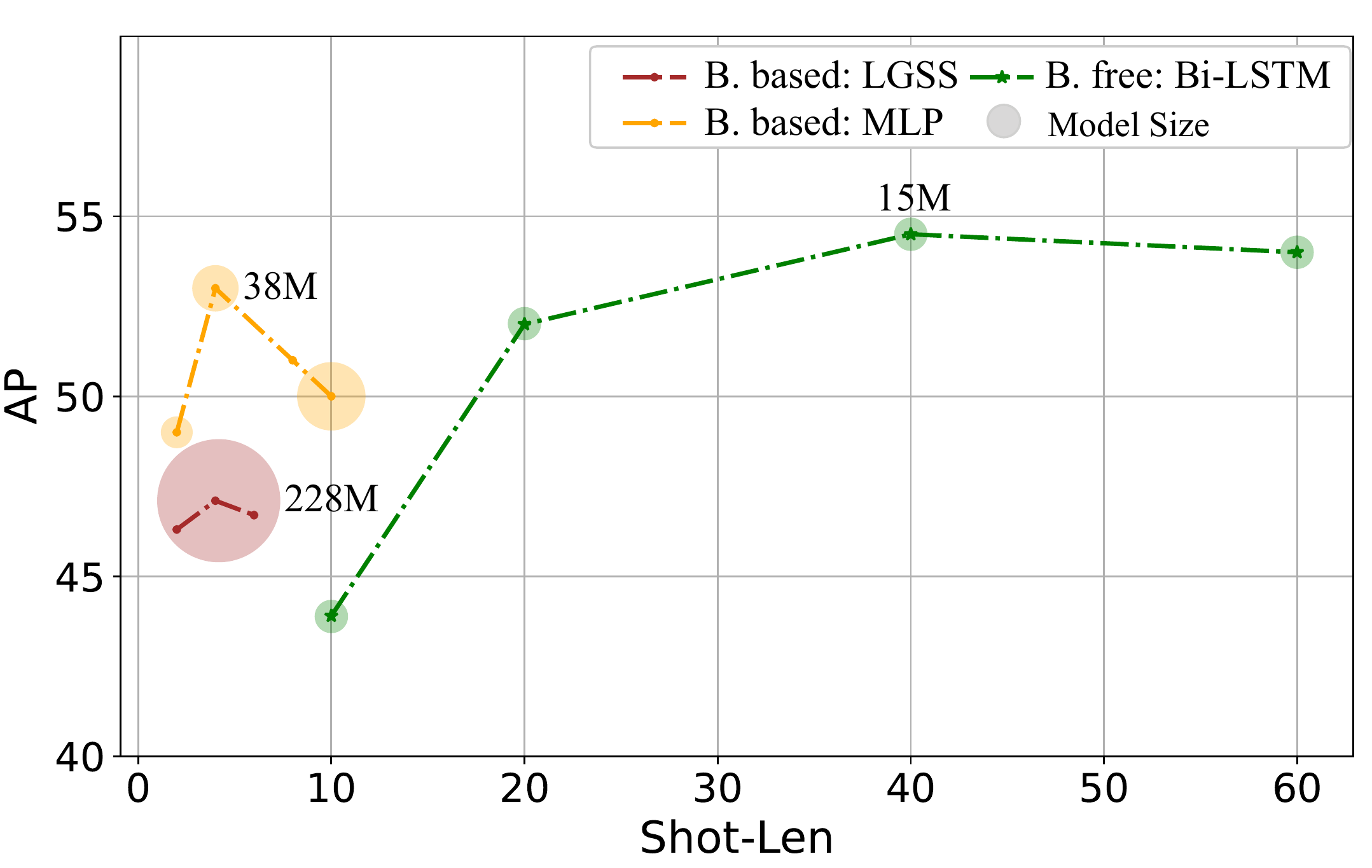}
  \caption{AP results on the MovieScene-318 dataset and model size of Boundary based and Boundary free models, where  \textbf{B.} stands for \textbf{Boundary} and $Shot$-$Len$ represents the length of shots processed within a batch.}
  \label{fig:boud}
\end{figure}

\begin{figure*}
  \centering
   \includegraphics[width=0.80\linewidth]{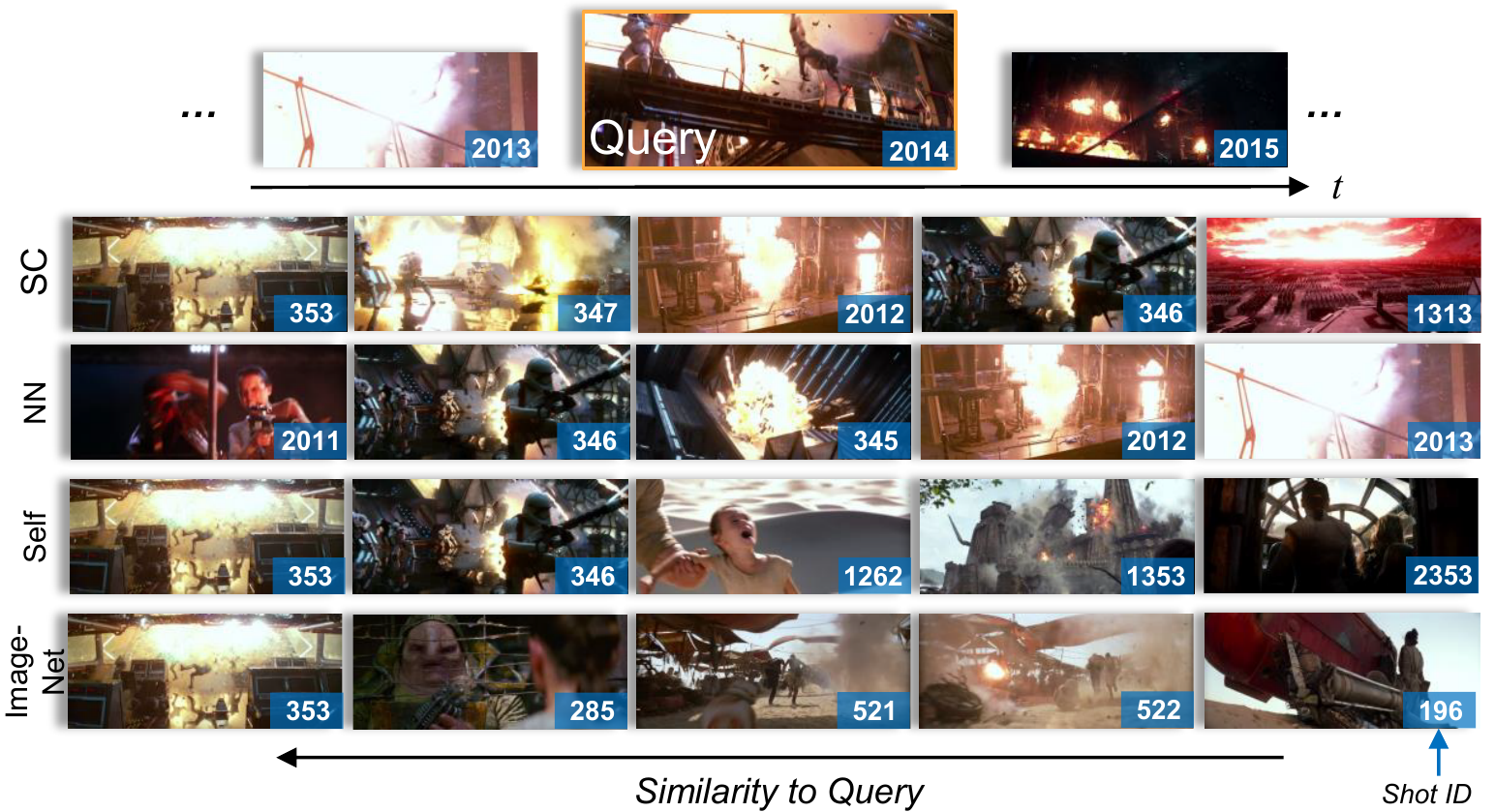}

   \caption{\textbf{The visualization results of shot retrieval}. 
   Compared with the other methods, the results of \textbf{\textit{SC}} appear more consistent in terms of the the semantic information, \textit{i.e.}, scenes with explosions.
   }
   \label{fig:retrieval}
\end{figure*}

\subsection{Results under MLP/Bi-LSTM protocols}

To study the superiority of the introduced evaluation protocol for the task of \textit{Video Scene Segmentation}, AP results together with the model size of Boundary based (\textit{i.e.,} LGSS \cite{rao2020local} and MLP \cite{chen2021shot} protocol) and Boundary free (\textit{i.e.,} introduced Bi-LSTM protocol) models are shown in Fig. \ref{fig:boud}.

As shown in Fig. \ref{fig:boud}, although performances of all models are associated with the length of the shots, \textit{i.e.,} $Shot$-$Len$, the Boundary based model achieves the best performance only when the $Shot$-$Len$ is relatively small (and the optimal $Shot$-$Len$ is less than the average number of shots per scene, \textit{i.e.} 12).
As $Shot$-$Len$ becomes larger, the performance decreases and the model size increases.
By contrast, the Boundary free model produces less inductive bias and takes the shot features as the unit of basic temporal input, hence, it is able to model representations of longer shots, while achieving better performances
when $Shot$-$Len$ takes a value in the approximate range and a model with the same size is employed.

\subsection{Visualization}

\subsubsection{Shot Retrieval}

An additional result of Shot Retrieval, which is introduced in the Sec. \textcolor{red}{4.4} of the main body of this work, is given in the Fig. \ref{fig:retrieval}.

\subsubsection{Scene Boundary}

\begin{figure*}
  \centering
\includegraphics[width=.95\textwidth]{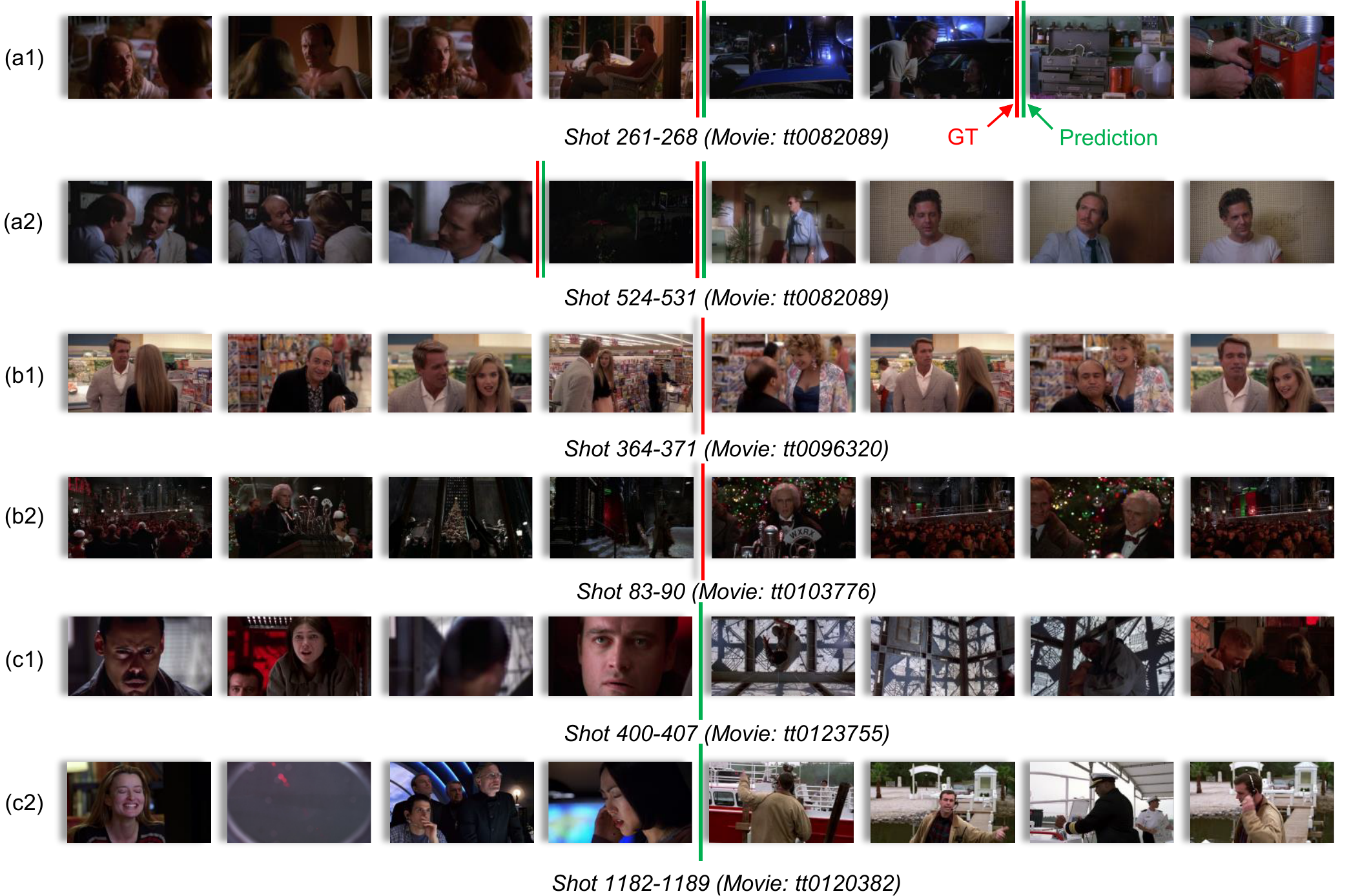}
  \caption{The ground truth (GT) and prediction scene boundaries are presented in this figure, where the middle frame of each shot is visualized. Fig. (a1)/(a2), Fig. (b1)/(b2) and Fig. (c1)/(c2) show the prediction cases of true positive, false negative and false positive, respectively.}
  \label{fig:pred}
\end{figure*}

To study the practical performance of our approach for the \textit{Video Scene Segmentation} task, we visualize the GT/Prediction scene boundaries in Fig. \ref{fig:pred}. For simple scenes in Fig. \ref{fig:pred} (a1)/(a2), the proposed method easily identifies these scenes and gives correct predictions of the scene boundary. 
As shown in Fig. \ref{fig:pred} (b1)/(b2)/(c1)/(c2), there are also bad cases where the proposed method fails to distinguish between the segmentation points of a shot and a scene, and for these cases, it may be confusing to identify whether these shots belong to the same scene or not, from the visual modality. 

\newpage
{\small
\bibliographystyle{ieee_fullname}
\bibliography{supplement}
}